\documentclass[10pt,journal,compsoc]{IEEEtran}

\ifCLASSOPTIONcompsoc
  \usepackage[nocompress]{cite}
\else
  \usepackage{cite}
\fi
\ifCLASSINFOpdf
\else
\fi
\usepackage{epsfig}
\usepackage{graphicx}
\usepackage{amsmath}
\usepackage{amssymb}
\usepackage{multirow}
\usepackage{tabularx}
\usepackage{booktabs}
\usepackage{mathtools}
\usepackage{enumerate}
\usepackage{xcolor}
\usepackage{booktabs}
\usepackage{color, colortbl}
\usepackage{bigstrut}
\usepackage{floatrow}
\usepackage{caption}
\usepackage{arydshln}
\usepackage{subfig}
\usepackage[export]{adjustbox}
\usepackage{algorithm}
\usepackage{algpseudocode}
\usepackage{pifont}
\newcommand{\cmark}{\ding{51}}%
\begin{document}
%
\title{Illuminating Salient Contributions in Neuron Activation with Attribution Equilibrium}
%
%
%
%

\author{Woo-Jeoung Nam, and~Seong-Whan Lee,~\IEEEmembership{Fellow,~IEEE}
\IEEEcompsocitemizethanks{\IEEEcompsocthanksitem W.-J. Nam is with the School of Computer Science and Engineering, Kyungpook National University(KNU), 80 Daehak-ro, Buk-gu, Daegu 41566, Republic of Korea. e-mail: nwj0612@knu.ac.kr.\protect\\
\IEEEcompsocthanksitem S.-W. Lee is with the Department of Artificial Intelligence, Korea University, Anam-dong, Seongbuk-ku, Seoul 02841, Republic of Korea. e-mail: sw.lee@korea.ac.kr.}
\thanks{Manuscript received April 2, 2022; revised xx xx, 2022.\\
Corresponding author: Seong-Whan Lee}}

%
%

\markboth{Journal of \LaTeX\ Class Files,~Vol.~xx, No.~x, August~20xx}%
{Shell \MakeLowercase{\textit{et al.}}: Bare Advanced Demo of IEEEtran.cls for IEEE Computer Society Journals}
%



\IEEEtitleabstractindextext{%
\begin{abstract}

With the remarkable success of deep neural networks, there is a growing interest in research aimed at providing clear interpretations of their decision-making processes. In this paper, we introduce Attribution Equilibrium, a novel method to decompose output predictions into fine-grained attributions, balancing positive and negative relevance for clearer visualization of the evidence behind a network decision. We carefully analyze conventional approaches to decision explanation and present a different perspective on the conservation of evidence. We define the evidence as a gap between positive and negative influences among gradient-derived initial contribution maps. Then, we incorporate antagonistic elements and a user-defined criterion for the degree of positive attribution during propagation. Additionally, we consider the role of inactivated neurons in the propagation rule, thereby enhancing the discernment of less relevant elements such as the background. We conduct various assessments in a verified experimental environment with PASCAL VOC 2007, MS COCO 2014, and ImageNet datasets. The results demonstrate that our method outperforms existing attribution methods both qualitatively and quantitatively in identifying the key input features that influence model decisions.

\end{abstract}

}

\maketitle

\IEEEdisplaynontitleabstractindextext

%
\IEEEpeerreviewmaketitle

\ifCLASSOPTIONcompsoc
\IEEEraisesectionheading{\section{Introduction}\label{sec:introduction}}
\else
\section{Introduction}
\label{sec:introduction}
\fi

%
%
%
%
\IEEEPARstart{W}ith advancements in deep neural networks (DNNs) across various fields of computer science, many studies have aimed to improve the transparency of these networks' decision-making processes. The most common method for explaining DNN decisions is the assignment of attribution (also known as relevance), which aims to understand decisions by highlighting the most relevant input factors and characterizing them as the basis for those decisions.
These theoretical methods enhance confidence in model decisions and provide intuitive understanding, even in specialized fields like medicine~\cite{binder2021morphological, hofmann2022towards} and chemistry~\cite{keyl2023single}. They are also commonly utilized in other computer vision applications\cite{ahn2019weakly, Lee_2019_CVPR, huang2018weakly}, e.g., weakly supervised segmentation and detection, as the method to obtain the initial seeds.

To identify relevant input parts for network prediction, many studies have used modified backpropagation algorithms~\cite{bach2015pixel,kindermans2017patternnet, montavon2017explaining, zhang2018top, nam2020relative, ancona2018towards, gur2021visualization, lee2021relevance, montavon2019layer}. These methods determine the importance of neurons across different layers by propagating output logits backward through the network. Each method interprets and visualizes network decisions based on its unique propagation rules, assigning relevance to the activated neurons according to their contributions. However, some methods suffer from class-agnostic, scattered attributions to irrelevant factors, and ambiguous criteria for interpretation that largely rely on human judgment.

A conservation rule, derived from layer-wise relevance propagation\cite{bach2015pixel}, maintains output values during backward propagation as evidence for a decision. This principle has become central to many attribution-based methods. Most methods normalize relevance between activated neurons in each layer, distributing it according to their contribution. Particularly in scenarios where the target class remains unchanged in a single image, variations in the output node do not alter the strength of attributions unless the sign of the value changes. From the standpoint of interpreting the model's results, the explanation should focus on the actual decision made by the network, not on a hypothetical one that the model has not computed. However, end-users might be interested in observing changes in the heatmap based on node values or in examining attributions related to different classes of the input image. From the perspective that positive and negative attributions represent high and low contributions of neurons to network decisions, we focus on balancing these attributions in relation to the evidence of the decision. Balancing these influences allows for a clearer visualization of the relationship between evidence and explanation. This clear representation is a valuable tool for understanding the underlying mechanisms of deep networks.
\begin{figure*}[!t]
  \centering
  \includegraphics[width=1\linewidth]{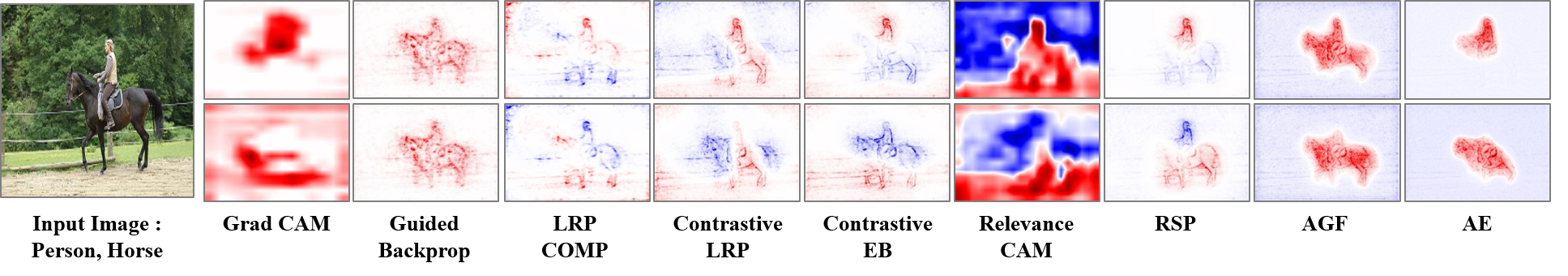}
    \caption{Our method aims to provide sophisticated visualizations of the significant input features related to network decision, effectively addressing challenges related to detail deficiency, class-discriminability, and clarity of object representation.}
  \label{fig:intro}
\end{figure*}

More formally, $f(I)=[f_1(I), \cdots, f_M(I)]$ is the output of a DNN with input $I=[I_1, \cdots I_N]$ and $M$ number of output neurons. The initial relevance is $R$, which has the same shape as the output tensor but only has a target ($t$) output logit, $f_t(I)$. Assume that $R$ is fully decomposed into the input features as follows: $R^t=[R_1^t, \cdots, R_N^t]$, and consider the opposite case, i.e., $-R$. Other attribution methods are based on finding features that contribute negatively to a decision, resulting in the same final attribution but with an opposite sign compared to the original case, $R$. In contrast, our approach aims to maintain the overall consistency of relevance by preserving the positive contributions of each neuron towards the evidence for a decision, while simultaneously altering irrelevant parts negatively as contradictory evidence. 

In this paper, we propose a new perspective for the evidence to network decision, allowing exploration of the fine-grained attributions. Fig.~\ref{fig:intro} illustrates the examples that summarize the characteristics of our method. The attributions in the input are fully decomposed from the network output with characteristics of i) being limited to a minimal yet highly relevant scope, denoting the most correlated parts to the network decision, ii) fine-grained localization to the detailed parts of main object, and iii) class-specific visualization and parameterizing the ratio of positive attribution. The main contributions of this study are as follows:
\begin{itemize}
\item We have developed an attribution method to decompose network predictions into significant input features, ensuring a clear and fine-grained visualization of the evidence behind decisions. We define the evidence leading to a decision based on the overwhelmingly positive attributions in the initial contribution maps derived from gradients and activation. Moreover, we properly allocate the relevance scores in line with preserving the salient properties, while denoting features related to contradictory evidence as negative scores. This approach achieves a balance between positive and negative attributions during propagation, leading to sophisticated and fine-grained upsampling of evidence.
\item 
As additional evaluations, we analyze the results of the pointing game (including GridPG), sanity check with model sensitivity, and outside–inside ratio to assess the quality of the attributions. To confirm the efficacy of understanding the models, we examine the performance in two scenarios (either only accurate or all labels). Consistent with our assumptions, the results demonstrate that our method possesses characteristics of objectness, class specificity, and provides clear descriptions of key input features.
\item 
We carefully report the unpredictable phenomenon of a DNN when assessing the attributions using perturbation-based evaluation metrics. Based on the fact that DNNs are vulnerable to widely dispersed noise and rapid inter-pixel variation, we introduce cases that inadvertently occur during perturbation-based evaluations. We confirm that assessments through region insertion with Gaussian filter are rather stable and discuss the necessity for additional research on attribution evaluation metrics.
\end{itemize}
\section{Related Studies}
Recently, many studies have aimed at increasing the interpretability of DNNs. As a method to analyze a DNN model itself, salient features are visualized by maximizing the activated neurons in intermediate layers~\cite{erhan2009visualizing} or generating saliency maps \cite{simonyan2013deep,zeiler2014visualizing,mahendran2016visualizing,zhou2016learning,dabkowski2017real,zhou2018interpreting}. A model agnostic method called LIME was proposed by \cite{ribeiro2016should}, which explains black-box models by locally approximating the decision boundary as simpler linear models.

A perturbation-based approach examines decision variations with gradual distortions in the network input. The large effect caused on the network output by the input perturbations leads to pathological perturbations, which induce adversarial effects ~\cite{fong2017interpretable}.  ~\cite{zeiler2014visualizing, petsiuk2018rise, Petsiuk_2021_CVPR} estimated the variations in the network output by applying occlusions with specified patterns such as random masking. ~\cite{fong2019understanding} established the concept of an extremal perturbation to comprehend the network decision with theoretical masking.

From the viewpoint of assigning attributions \cite{lim2021building,lee2021relevance,chockler2021explanations,singla2021understanding,jung2021towards}, ~\cite{bach2015pixel} first provided the concept of relevance and conservation for the evidence to the decision using several types of layer-wise relevance propagation (LRP) rules. ~\cite{montavon2017explaining, montavon2019layer} proposed deep Taylor decomposition (DTD) based on Taylor expansion in intermediate layers with a theoretical foundation. DTD decomposes the network decision into contributions of the input features. DeepLIFT \cite{shrikumar2017learning} focused on clarifying the differences in the contribution scores of neurons and their reference activation. \cite{ancona2018towards} studied attributions from a theoretical standpoint, formally proving the conditions of equivalence of previous methods. \cite{lundberg2017unified} introduced the Shapley value to unify and approximate existing explaining methods. By modeling with the probabilistic winner-take-all process, ~\cite{zhang2018top} proposed excitation backprop with top-down attention. ~\cite{nam2020relative} first presented an influence perspective to resolve the overlapping phenomenon of positive and negative contributions, resulting in clear separation of a target object and its irrelevant background. \cite{lapuschkin-ncomm19} examined misleading correlations among the objects in a network input and highlighted the necessity of understanding a network decision to reveal the ``clever Hans'' phenomena. However, methods based on LRP have limitations originating from the class-agnostic issue of DNN models, which results in same visualizations even when starting from different classes. ~\cite{zhang2018top} explored the cause of this issue as ``winner always wins,'' by which phenomenon highly activated neurons have the majority of the relevance during backward propagation. As an alternative, ~\cite{zhang2018top, gu2018understanding} presented contrastive concepts by propagating the relevance from the target in contrast to all other classes.

Some attribution approaches utilize the gradients with respect to the target based on the chain rule. Gradient-weighted class activation map (Grad-CAM)\cite{selvaraju2017grad} is the most well-known and commonly used method owing to its easy applicability and high performance in localizing primary objects. It generates class-specific activation maps by computing gradients with respect to target activation neurons in the feature extraction stage. Guided BackProp~\cite{springenberg2014striving} is based on gradient backpropagation considering only positive values and aims to clarify specific input features. Integrated gradients \cite{sundararajan2017axiomatic} utilize average partial derivatives of a network output to resolve the gradient saturation problem. SmoothGrad~\cite{smilkov2017smoothgrad} is a method for visualizing mean gradients with addition of a random Gaussian noise to an image. FullGrad\cite{srinivas2019full} computes the gradients of the input and bias of each layer and sums over the complete network to visualize an entire saliency map.

As an integrated method of relevance and gradients, AGF\cite{gur2021visualization} introduces attribution-guided factorization to extract class-specific attributions derived from input features and gradients. RSP\cite{nam2021interpreting} provides class-specific and intuitive visualizations by distinguishing hostile activations with respect to the target class.

\section{Evidence for Decision}
In this section, we introduce the mechanisms of general propagation rules and address different views on the evidence for a decision.
\subsection{Background}
LRP was first introduced by \cite{bach2015pixel}, which finds the highly contributing parts in an input, by decomposing output predictions in a backward manner. It is based on the conservation principle, which treats the output of a network as the evidence for a decision and maintains it in each layer during propagation. More formally, assuming that \(R_{j}^{(l)}\) denotes the relevance of a neuron \(j\) in a layer \(l\) and that \(R_{i}^{(l-1)}\) is the assigned relevance by the propagation rule in layer \(l-1\), the above-mentioned conservation takes the form,
\begin{equation}
    \sum_i R_{i}^{(l-1)} = \sum_j R_{j}^{(l)}.
    \label{eq:conservation}
\end{equation}
The forward process between layers $l-1$ and $l$, to which attributions are propagated, is denoted as $\mathcal{F}(x^{(l-1)},w^{(l-1,l)})$, where $x$ and $w$ represent a neuron and its weight, respectively. Here, we define the generic relevance propagation rule \cite{montavon2017methods} that operates in a backward manner with the normalization process based on the contributions of the neurons as follows:
\begin{equation}
\begin{aligned}
R_{i}^{(l-1)} & = \mathcal{J}(x_{i}^{(l-1)},w^{(l-1,l)},R_{j}^{(l)})\\
& = \sum_{j} \frac{x^{(l-1)}_{i}w^{(l-1,l)}_{i,j}}{\sum_{i^{\prime}}x^{(l-1)}_{i^{\prime}}w^{(l-1,l)}_{i^{\prime},j}}R_{j}^{(l)}.
\label{eq:prop}
\end{aligned}
\end{equation}
Here, $i$ and $j$ represent the indices of a neuron in layers $l-1$ and $l$, respectively, and Eq. \ref{eq:prop} preserves the conservation rule in Eq. \ref{eq:conservation}, which maintains the total relevance as constant during propagation.
To consider the positive and negative contributions of neurons, \cite{bach2015pixel} introduced a rule LRP-\(\alpha\beta\) that enforces the conservation principle. For intuitive understanding, we assume that the network is composed of ReLU activation functions, which means the sign of the feedforward value is determined by the sign of the corresponding weight. To maintain the total relevance, parameters are chosen such that \(\alpha - \beta = 1\). Each part of the relevance with positive and negative weights, marked as $w^+$ and $w^-$, respectively, is multiplied by $\alpha$ ($\beta$) and related to positive (negative) activations as positive (negative) relevance.
\begin{equation}
\begin{aligned}
    R_{i}^{(l-1)} & =  \alpha*\mathcal{J}(x_{i}^{(l-1)},w^{+(l-1,l)},R_{j}^{(l)})\\
    & -\beta*\mathcal{J}(x_{i}^{(l-1)},w^{-(l-1,l)},R_{j}^{(l)}).
    \label{eq:ablrp}
\end{aligned}
\end{equation}
The obtained attributions are mapped to the pixels of the input image and shown as a heatmap after relevance propagation is completed from the last to the start layer.
Composite-LRP\cite{montavon2019layer} combines various versions: LRP-$\{0, \epsilon, \gamma\}$ tailored to each layer within the model, providing clearer and enhanced explanations.

\begin{figure}[t!]
  \centering
  \includegraphics[width=1\linewidth]{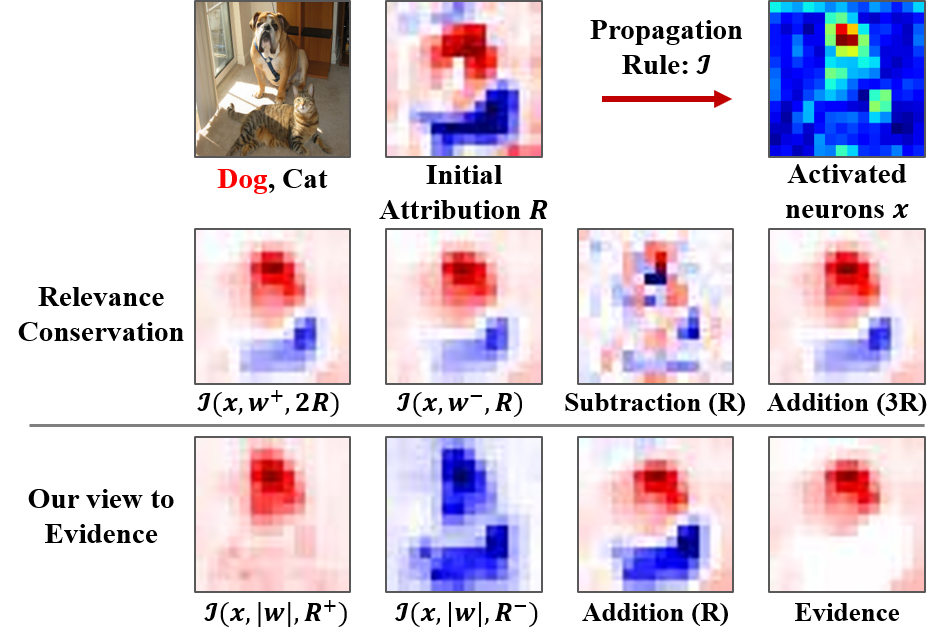}
    \caption{Images in the First row represent the input, initial attribution $R$, and neurons to be assigned the relevance, respectively. The second row shows the issue when propagating the relevance according to the actual contributed value. The third row shows our view of influence on propagation and the criterion of evidence that overwhelms the negative influence. Detailed explanation is given in Section 3.2.}
  \label{fig:paradox}
\end{figure}
\subsection{Discussion to Conservation Rule}
First, we address the consistency of assigned attributions when varying the evidence. In previous studies, the output logit before passing through the final classifier layer, such as softmax, was regarded as the evidence. Since this input relevance takes the form of a one-hot vector, increasing or decreasing the target node value does not affect the direction (positive or negative) of an attribution itself. This approach, intended to interpret the decisions made by the model itself, is fundamentally sound. If the model were to produce a different output logit, then the explanation would accordingly change. However, end-users might wish to parameterize the explanation in relation to the evidence to identify which areas within the input are more or less important, based on the degree of variation. If the propagation rule is based on normalization (e.g., Eq.~\ref{eq:prop}) and the logit value is adjusted by a user to be negative, the strength of all final attributions is preserved, and only a sign reversal occurs.

The second issue concerns the conservation rule for propagating the initial attribution (e.g., GradCam) generated from the classification stage. Fig. \ref{fig:paradox} illustrates motivational examples for intuitive understanding. The input image containing a dog and a cat is correctly classified by the DNN. The initial attribution (described in Section 4.1), $R$, is generated by utilizing the gradient and activated neurons relative to the dog label. This attribution is normalized so that the positive and negative values have the same total absolute values, i.e., $\sum{R}=0$. For visualization, we apply a channel-wise summation in a certain layer. Although fundamentally different from LRP, we apply this relevance $R$ to the neurons in the next layer using the general propagation rule in Eq.\ref{eq:prop} for testing purposes. The second row in the figure shows the partial results based on Eq.\ref{eq:ablrp} when $(\alpha=2,\beta=0)$ and $(\alpha=0,\beta=1)$ before subtraction. Generally, because the value propagated with the negative weight in a forward pass is a negative contribution to a decision, it might seem intuitive to assign negative relevance to the neurons corresponding to the dog. However, due to the dependence on neuron activation values, the results are not significantly different from those based on positive weights. When they are subtracted from each other as per Eq.\ref{eq:ablrp}, the conservation rule is preserved, and the differences within the attributions maintain the same direction. If the two partitions are added, although the directions are maintained, the conflict between positive and negative attributions is inevitable.

We first identified this problem in our previous study \cite{nam2020relative} and approached it from the perspective of the neuron's influence. This approach indicates that highly activated neurons play important roles in the network decision in both positive and negative ways, necessitating a dominant share in the relevance allocation. Further refined here, we intend to propagate the input relevance separately, to maintain the positive and negative directions \cite{nam2021interpreting}. The third row in the figure illustrates our view with respect to the evidence for the dog. The first and second images visualize the positive and negative attributions from each section of the initial attribution, respectively. Although the cat-related features barely contribute to the those of the target dog, they negatively influence the primary object through negative weights. We consider the gaps where positive attributions outweigh the negatives in the case of equally limited assignments as the evidence for the network decision. Users should be able to observe the variations in the final assigned attributions by increasing or decreasing this evidence as desired.

Fig.~\ref{fig:intro} shows the results of prominent explaining methods based on a modified backpropagation algorithm with their purposes. Although Grad-CAM is well known for its remarkable localization performance with respect to a target, many losses are caused in the description of the details of pixel-level granularity by the skipping of the interpretation of the feature extraction stages. Other methods can fully decompose a prediction in a backward manner, including in the feature extraction stage. 
Although Guided BackProp represents attributions at a pixel level, there is no visual difference in the attributions per class. Methods with a contrastive perspective, including Relevance-CAM and LRP-COMP, provide class-specific attributions according to the target class. However, positive or negative relevance is disseminated in the background or in other sections unrelated to its origin. Because this approach yields a relatively larger activated area than other classes, irrelevant areas could receive positive or negative relevance scores.

In this study, we mainly explore the most significant attributions corresponding to a target. Although relative sectional propagation (RSP) and attribution-guided factorization (AGF) provide reasonable explanations of a decision with intuitive visualization, we aim to identify the intensive input features that largely influence a network, instead of targeting semantic segmentation.

\section{Proposed Method}
Our method for backward decomposition of the network output involves two main stages: (i) acquiring initial contribution maps from the gradients of the target class, and (ii) balancing the positive and negative attributions in line with their contribution to evidence.
\begin{figure}[t!]
  \centering
  \includegraphics[width=1\linewidth]{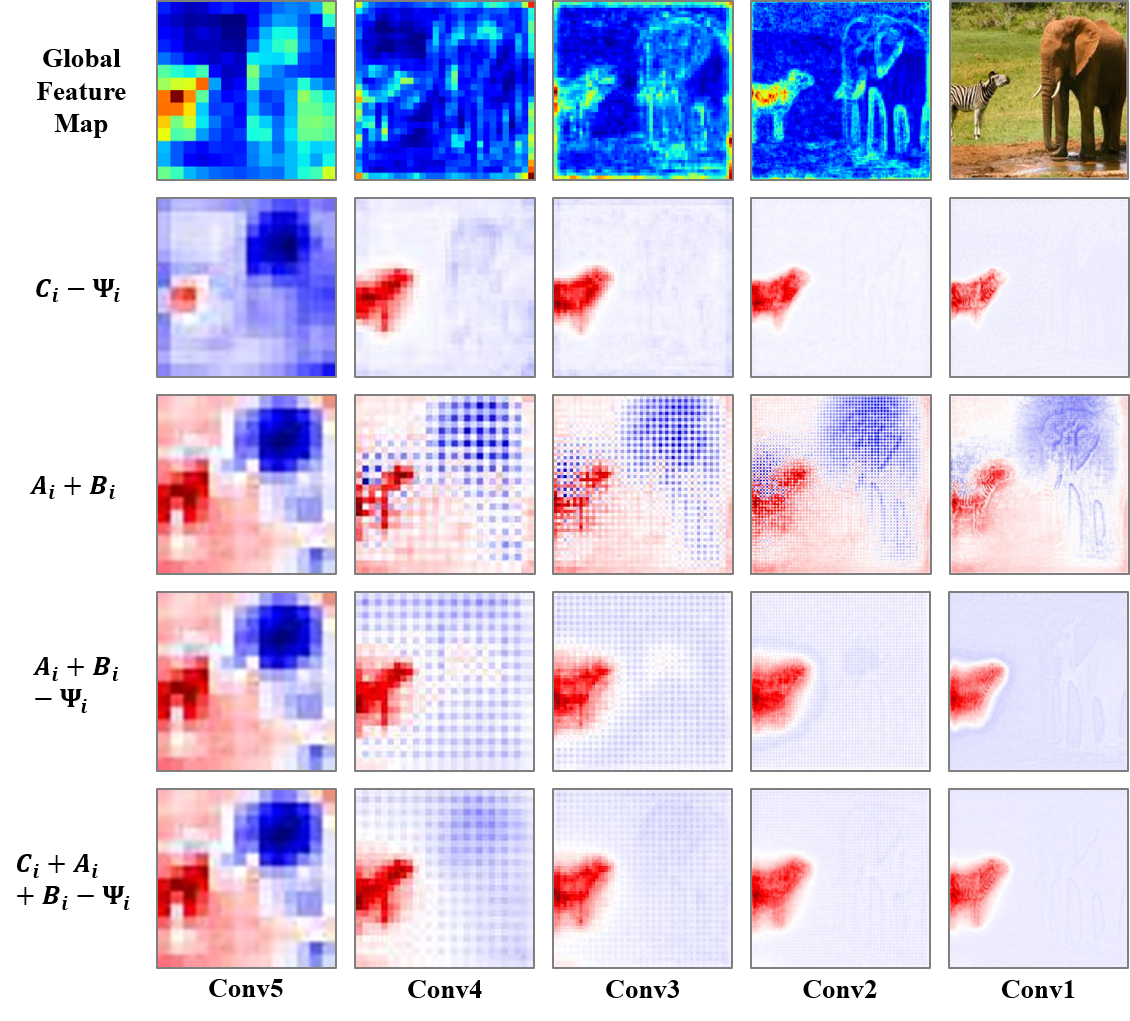}
    \caption{Difference between channel attributions of intermediate layers with/without considering activation properties. First row shows global input features of intermediate layers. Each row below visualizes channel-wise sum of attributions based on variants.}
  \label{fig:stage}
\end{figure}
\subsection{Initial Contribution Maps}
This section introduces a method to acquire gradient activation maps for a target class. We denote the layers sequentially from the intermediate layer $k$ to the input layer $0$. Let $y$ and $t$ represent the values of the network output and the target class node, respectively. Using the chain rule, the gradients of the input features in intermediate layers $k$ and $k-1$ can be obtained as follows:
\begin{align}
    \begin{gathered}
        \frac{\partial y^{(t)}}{\partial x^{(k-1)}_i} = \sum_{j}\frac{\partial y^{(t)}}{\partial x^{(k)}_j}\frac{\partial x^{(k)}_j}{\partial x^{(k-1)}_i}
    \end{gathered}.
    \label{eq:chain}
\end{align}
We backpropagate the gradient of \(y\) to a user-selected intermediate layer $l$ (input features prior to the averaging pooling layer in our study) using the chain rule. This gradient serves as a channel-wise neuron importance.
\begin{align}
    \begin{gathered}
        G^{(l)}_{c,h,w} = \lambda * x^{l}_{c,h,w} * \frac{1}{Z}\sum_{h}\sum_{w}\frac{\partial y^{(t)}}{\partial x^{(l)}_{c,h,w}}
    \end{gathered}.
    \label{eq:cam}
\end{align}
$x^{(l)}_{c,h,w}$ represents each neuron in the $c^{th}$ feature map with a matrix of size $Z=H*W$ and indexed by pixel $(h,w)$. The propagated gradients are global average-pooled to serve as neuron importance weights with respect to the activated neurons. $\lambda$ denotes a normalization factor that divides the gradient activation map by the absolute value of the total sum, $|\sum_{c,h,w}G^{(l)}_{c,h,w}|$, for computational efficiency. Eq.~\ref{eq:cam} used for computing a gradient activation map is similar to Grad-CAM except elimination of the negative values and the last linear combination between the feature map and the partial linearization. While there are improved version of Grad-CAM, such as Score-CAM\cite{wang2020score} and Grad-CAM++\cite{chattopadhay2018grad}, they do not offer significant advantages over our approach in terms of the initial input attributions.

In our gradient activation map, positive and negative values are determined by the gradient with respect to the target class, while unrelated objects or the background are assigned values close to zero or negative. Here, under the assumption that the absolute sums of the positive and negative attributions are equal, we identify the dominant positive attributions over the negative ones as the evidence for a decision.

\subsection{Propagation rules of Attribution Equilibrium}
Assuming that the evidence in the initial attributions indicates a purely positive contribution, the negative influence of irrelevant objects or the background represents contradictory evidence for the target. Before backward propagating the initial attributions from neuron $j$ in layer $l$ to neuron $i$ in layer $l-1$, we adjust the sum of all positive and negative neuron values to be the same absolute value, without changing the sign of each neuron. We denote the positive and negative sections of the initial attributions as $\mathcal{P}^{(l)}$ and $\mathcal{N}^{(l)}$, respectively, and modify them as follows:
\begin{equation}
\mathcal{P}_{j}^{\prime(l)}=\mathcal{P}_{j}^{(l)}*\frac{\sum_{j}|G^{(l)}_{j}|}{\sum_{j}{\mathcal{P}^{(l)}_{j}}}, \quad
\mathcal{N}_{j}^{\prime(l)}=\mathcal{N}_{j}^{(l)}*\frac{-\sum_{j}|G^{(l)}_{j}|}{\sum_{j}{\mathcal{N}^{(l)}_{j}}},
\label{eq:3}
\end{equation}
\begin{equation}
R_{j}^{(l)} = \gamma*\mathcal{P}_{j}^{\prime(l)} + \mathcal{N}_{j}^{\prime(l)},\quad 1\leq\gamma\leq2,
\label{eq:4}
\end{equation}
where $\gamma$ is a user preference parameter that determines the degree of evidence to be preserved during propagation. Using this attribution, we compute the contributions of the neurons in layer $l-1$ to each section of the positive and negative attributions from an influence perspective \cite{nam2020relative}. More specifically, the influencing perspective is that activated neurons with higher values are more critical in both positive and negative manners, resulting in more dependence on their values. Therefore, for these neurons, the relevance should be allocated depending on the degree of influence, regardless of the positive or negative direction. We utilize both positive and negative weights by casting the absolute values. For the positive and negative sections, each contribution is computed as follows:
\begin{algorithm}[t]
\caption{Propagation with Attribution Equilibrium}
\begin{algorithmic}[1]
\Require Input $I$, Model: $F$, Target: $t$, Neurons in layer $l$:
$x^{(l)}$, \quad$l$ starts from end to start layer
\State $y\leftarrow F(I)$  \Comment{Forward pass}
\State $\hat{y}\leftarrow y^t\,*\,$Onehot$(t)$  \Comment{One-hot encoding}
\For{layer $l$ in Classification stages}
    \State $\nabla x^{(l)}\leftarrow \frac{\partial y^{t}}{\partial x^{(l)}}$    \Comment{Gradient propagation by chain rule}
    \If{layer $l$ in Average Pooling}
    \State $G^{(l)}\leftarrow x^{(l)}\, *\, $GAP$(\nabla x^{(l)})$
    \State $R^{(l)}\leftarrow \lambda*G^{(l)}$
    \Comment{Initial contribution map}
    \EndIf
\EndFor
\For{layer $l$ in Feature Extraction stages}
    \State Let $k=l-1$, $\theta:$ weights, 
    \State $R^{(l)}:=\gamma*\mathcal{P}^{(l)}+\mathcal{N}^{(l)}\leftarrow \mathcal{M}(R^{(l)},\gamma)$
    \State $\mathbf{C}^{(k)} \leftarrow \mathcal{J}(x^{(k)},|\theta|,\gamma*\mathcal{P}^{(l)}) +\mathcal{J}(x^{(k)},|\theta|,\mathcal{N}^{(l)})$
    \State $\mathbf{A}^{(k)} \leftarrow \mathcal{J}(\alpha^{(k)},|\theta|,\mathcal{P}^{(l)})$
    \State $\mathbf{U}^{(k)} \leftarrow \mathcal{J}(\beta^{(k)},|\theta|,\mathcal{N}^{(l)})$
    \State $\mathbf{R}_{i}^{(k)} = \mathbf{C}_{i}^{(k)} + \mathbf{A}_{i}^{(k)} + \mathbf{U}_{i}^{(k)} - \Psi_{i}^{(k)}$ \Comment{Update}
\EndFor
\label{alg:1}
\end{algorithmic}
\end{algorithm}
\begin{equation}
\begin{aligned}
\mathbf{C}_{i}^{(l-1)} = &  \mathcal{J}(x_{i}^{(l-1)},|w^{(l-1,l)}|,\gamma*\mathcal{P}_{j}^{\prime(l)})\\
&+\mathcal{J}(x_{i}^{(l-1)},|w^{(l-1,l)}|,\mathcal{N}_{j}^{\prime(l)}).
\label{eq:5}
\end{aligned}
\end{equation}
\begin{figure*}[t!]
  \centering
  \includegraphics[width=1\linewidth]{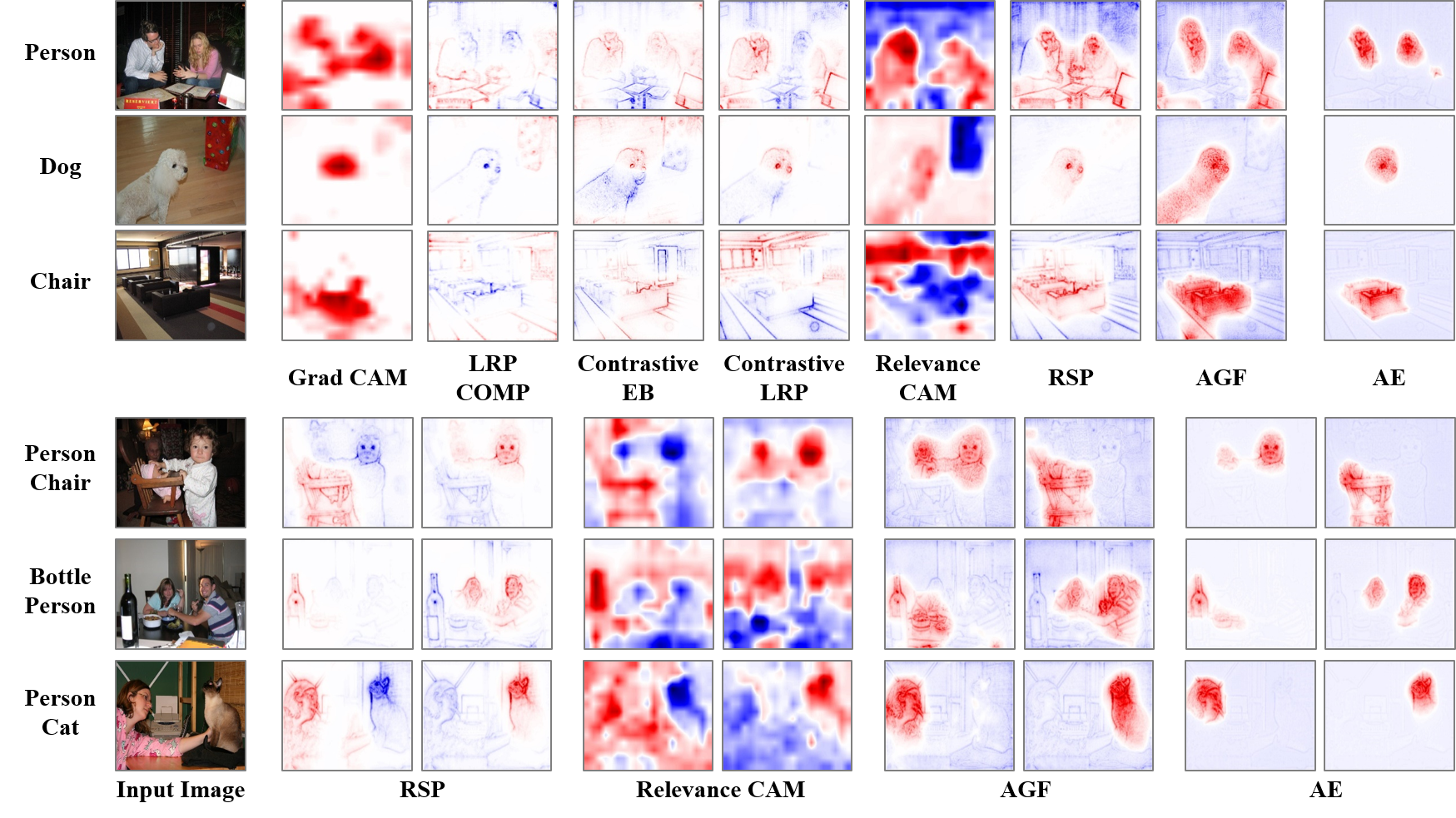}
    \caption{Comparison of conventional and proposed attribution methods applied to VGG-16 on PASCAL VOC dataset. Class names on left side represent predicted labels of input image. Upper and bottom groups show attributions for predictions of single- and multi-label images. Red and blue colors represent positive and negative values, respectively.}
  \label{fig:comp}
\end{figure*}

Each positive and negative attribution is backpropagated to an activated neuron in previous layer $l-1$ based on the neuron contribution, maintaining the overall relevance sum as $\gamma-1$. We revisit two points that affect allocation of precise attribution: i) the gap between the focused features among the first and end layers and ii) the role of an inactivated neuron. It is well known that the extracted features from each layer are different: a deep layer implies an abstract feature. Fig.\ref{fig:stage} shows examples of the variations in the input features in intermediate layers. Because of the dependence of the attribution methods on the value of an activated neuron, even if a slightly accurate relevance is initially assigned in the end stage of the network, the relevance could be biased in the initial stage owing to the presence of activated neurons with high values in irrelevant areas such as edge or background of an image. If the DNN's decision is influenced by watermarks or the edges of objects, these become important input features and should be emphasized (described in Fig.~\ref{fig:water}). Therefore, it is necessary to guide propagation to reflect the initial properties, including feature importance, which are directly related to network decisions. First, we define the mask of the neurons in each layer based on their activation as follows:
\begin{equation}
\begin{gathered}
\alpha_{i}^{(l)} = 
\begin{cases}
1  &,    \text{ $x_{i}^{(l)}>0$ } \\
0  &,    \text{ $x_{i}^{(l)}\leq0$ } 
\end{cases},\quad
\beta_{i}^{(l)} = 
\begin{cases}
0  &,    \text{ $x_{i}^{(l)}>0$ } \\
1  &,    \text{ $x_{i}^{(l)}\leq0$ } 
\end{cases}.\\
\end{gathered}
\label{eq:6}
\end{equation}
Subsequently, we design to propagate the positive attributions with respect to the target to be assigned to the activated neurons. The mask of the activated neurons receives positive relevance corresponding to the influence, regardless to the values of the neurons.
\begin{equation}
\begin{gathered}
\mathbf{A}_{i}^{(l-1)} = \mathcal{J}(\alpha_{i}^{(l-1)},|w^{(l-1,l)}|,\mathcal{P}_{j}^{\prime(l)}).
\end{gathered}
\label{eq:7}
\end{equation}
This propagation reduces the sensitivity of the attributions to local textures and preventing misguidance to local features such as edges and watermarks.

As the third variant, we consider the role of the inactivated neurons. In a forward pass, the neurons most irrelevant to the output are the inactivated neurons, due to the disconnection of value transfer. Therefore, we assign the negative attributions, which do not contribute to the decision, to the inactivated neurons by the following rule:
\begin{equation}
\begin{gathered}
\mathbf{U}_{i}^{(l-1)} = \mathcal{J}(\beta_{i}^{(l-1)},|w^{(l-1,l)}|,\mathcal{N}_{j}^{\prime(l)}).
\end{gathered}
\label{eq:8}
\end{equation}
The all-ones tensor corresponding to an inactivated neuron receives negative relevance corresponding to the influence of the weight. In our previous study~\cite{nam2020relative, nam2021interpreting}, uniform shifting presented excellent performance in separating the foreground and the background by changing the irrelevant attributions with relevance scores near zero into negative attributions. This allows the essential attributions to remain positive, with the irrelevant and hostile attributions becoming negative. We uniformly divide the absolute sum of the initial gradient activation maps, $\tau=|\sum{G}|$, by the number of activated neurons and subtract it from the summation of the variant equations as expressed below. For each iteration, the total uniform shifting value remains consistent.
\begin{equation}
\Psi_{i}^{(l-1)}=\frac{\tau}{\sum_i{\alpha_{i}^{(l-1)}}},
\label{eq:9}
\end{equation}
\begin{equation}
\mathbf{R}_{i}^{(l-1)} = 
\mathbf{C}_{i}^{(l-1)} + \mathbf{A}_{i}^{(l-1)} + \mathbf{U}_{i}^{(l-1)} - \Psi_{i}^{(l-1)}.
\label{eq:10}
\end{equation}
Relatively unimportant attributions, which are approximately zero, are converted into negative attributions during the propagation procedure. When progressing to the next layer, neurons positively contributed to these negative attributions are assigned the corresponding negative relevance scores, thereby the irrelevant attributions, e.g., the background, have negative relevance scores in the final output.
The sum of the attributions, $R_{i}^{(l-1)}$, is maintained as $(\gamma-2)*\tau$ in each iteration, except for the $1x1$ convolutional layer, which modulates the number of channels. In this case, $\gamma$ is set as $1$ and we do not utilize uniform shifting to prevent distortions along the channel map. In case of restricted relevance scores that assign contradictory evidence as negative, the propagated relevance explores the salient attributions by repeating the process in Eq.~\ref{eq:10} in each layer. This procedure is repeated until the first layer, $l=1$, of the model. For the final propagation to the input layer, we utilize \(Z^{\beta}\) rule\cite{bach2015pixel}, which is commonly used for the last propagation, resulting in clear visualization without interference of the attribution priorities. Algorithmic pseudo-code 1 describes the complete propagation rule.
\begin{figure*}[t!]
  \centering
  \includegraphics[width=1\linewidth]{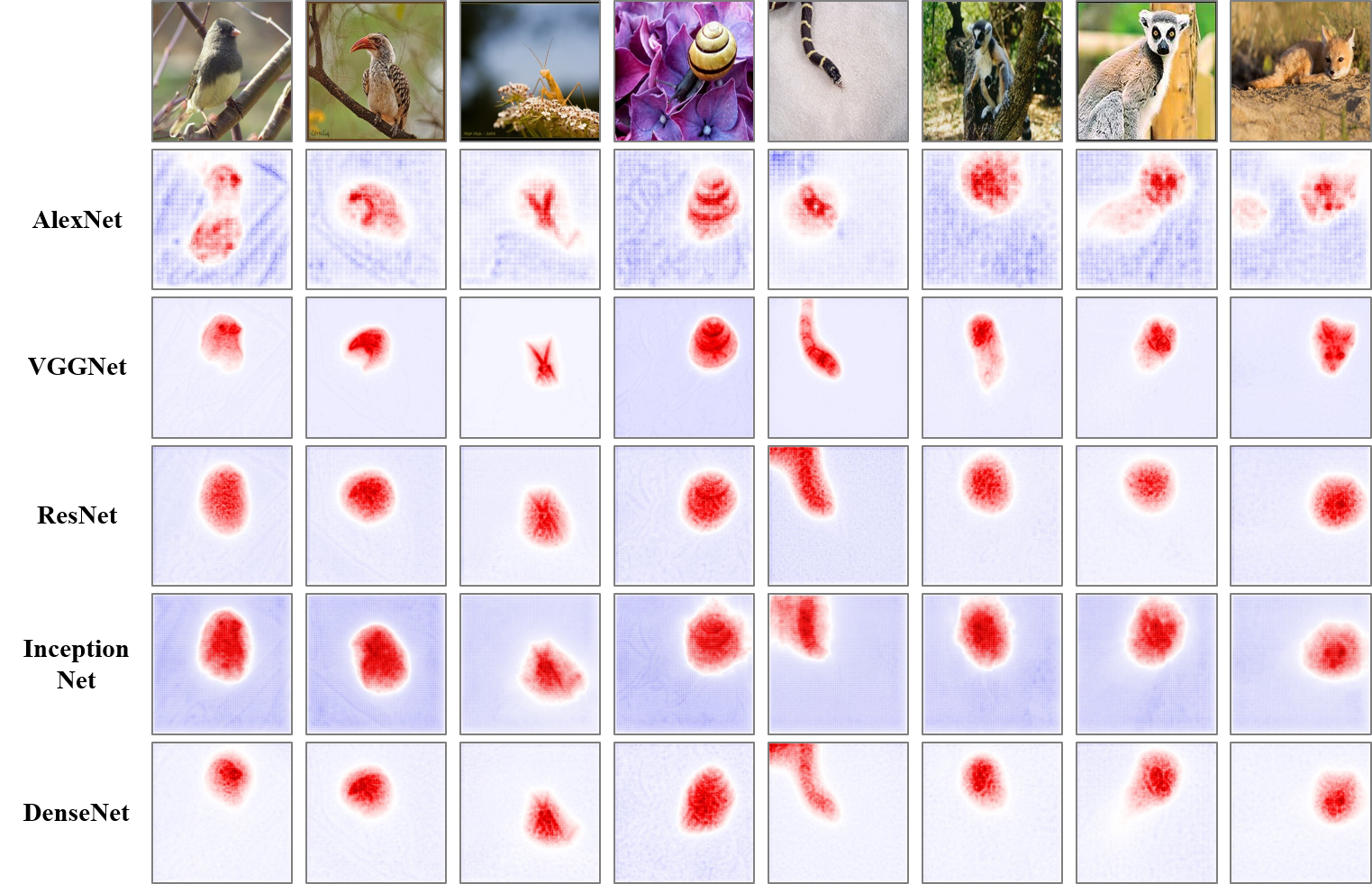}
    \caption{Applications of our method to various models: AlexNet, VGG-16, ResNet-50, Inception-V3, and DenseNet-121 on ImageNet validation dataset.}
  \label{fig:model}
\end{figure*}

\section{Experimental Evaluation}
\subsection{Implementation Details}
We mainly utilize commonly adopted CNN architectures, VGGNet and ResNet. We confirm that our method properly performs on the extended version of each network, i.e., VGG-19 and ResNet-121. Each model is trained on publicly accessible datasets Pascal VOC 2007 \cite{everingham2010pascal} and MS COCO 2014 \cite{lin2014microsoft}. For reasonable experiments, we adopt the trained models available online with the TorchRay package \cite{fong2019understanding}. We also present the evaluation of additional CNN models---DenseNet-121\cite{huang2017densely}, Inception-V3\cite{szegedy2016rethinking}, and AlexNet\cite{krizhevsky2012imagenet}---which are trained on the ImageNet 2012 dataset\cite{ILSVRC15} and publicly accessible. The attributions obtained using each method are visualized by seismic colors, with red and blue colors denoting positive and negative values, respectively. We follow the official implementations of other explaining methods to prevent erroneous reports. All conditions in the experiments are the same except the setting of the target saliency layer is according to the original design of each method. Each type of evaluation is described in the subsequent sections.
\begin{table*}[t]
\RawFloats
  \centering
    \resizebox{1\columnwidth}{!}{
    \large
    \begin{tabular}{cc|cc|cc|cc|cc|cc|cc|cc|cc}
        \specialrule{2.0pt}{1pt}{1pt}
     & &  \multicolumn{8}{c|}{\textbf{PASCAL VOC 2007}} & \multicolumn{8}{c}{\textbf{COCO 2014}} \bigstrut[b]\\
          & &\multicolumn{4}{c|}{VGG-16}  &\multicolumn{4}{c|}{ResNet-50} &\multicolumn{4}{c|}{VGG-16}  &\multicolumn{4}{c}{ResNet-50}  \bigstrut[b]\\
          & &\multicolumn{2}{c}{ALL}&\multicolumn{2}{c|}{DIF}  &\multicolumn{2}{c}{ALL}&\multicolumn{2}{c|}{DIF} &\multicolumn{2}{c}{ALL}&\multicolumn{2}{c|}{DIF}  &\multicolumn{2}{c}{ALL}&\multicolumn{2}{c}{DIF}  \bigstrut[b]\\
    METHOD &T& PG & PR  & PG & PR & PG & PR  & PG & PR  & PG & PR  & PG & PR  & PG & PR  & PG & PR\\
    \hline
    \multirow{2}[2]{*}{Grad-CAM} & L
    & .866  & .698
    & .740  & .677
    & .903  & .651
    & .823  & .641
    & .542  & .711
    & .490  & .688
    & .573  & .677
    & .523  & .668      \bigstrut[t]\\
          
    & P 
    & .945  & .772
    & .924  & .733
    & .953  & .736
    & .932  & .715
    & .727  & .741
    & .689  & .711
    & .705  & .723
    & .674  & .702 \bigstrut[b]\\
    \hdashline
    
    \multirow{1}[2]{*}{Guided} & L
    & .758  & --
    & .530  & --
    & .771  & --
    & .594  & --
    & .365  & --
    & .288  & --
    & .410  & --
    & .340  & --       \bigstrut[t]\\
          
    \multirow{1}[2]{*}{BackProp}& P
    & .880  & --
    & .784  & --
    & .857  & --
    & .756  & --
    & .600  & --
    & .536  & --
    & .573  & --
    & .519  & -- \bigstrut[b]\\
    \hdashline
    \multirow{1}[2]{*}{c*Exitation} & L 
    & .766  & .474
    & .634  & .468
    & .857  & .404
    & .741  & .411
    & .472  & .466
    & .417  & .453
    & .536  & .387
    & .485  & .397      \bigstrut[t]\\
          
    \multirow{1}[2]{*}{BackProp}& P 
    & .856 & .456
    & .784  & .434
    & .945  & .398
    & .887  & .407
    & .659  & .458
    & .620  & .433
    & .671  & .382
    & .636  & .388 \bigstrut[b]\\
    \hdashline
    \multirow{2}[2]{*}{LRP-COMP} & L
    & .702  & .473
    & .502  & .444
    & .843  & .355
    & .771  & .323
    & .411  & .466
    & .364  & .465
    & .515  & .367
    & .446  & .371       \bigstrut[t]\\
          
    & P  
    & .831  & .468
    & .755  & .448
    & .897  & .343
    & .851  & .331
    & .618  & .455
    & .574  & .432
    & .606  & .344
    & .581  & .348  \bigstrut[b]\\
    \hdashline
    \multirow{2}[2]{*}{c*LRP} & L
    & .719  & .461
    & .521  & .464
    & .852  & .341
    & .779  & .333
    & .434  & .454
    & .388  & .451
    & .533  & .361
    & .479  & .367       \bigstrut[t]\\
          
    & P  
    & .851  & .451
    & .773  & .418
    & .903  & .323
    & .871  & .311
    & .643  & .447
    & .591  & .441
    & .638  & .358
    & .611  & .354  \bigstrut[b]\\
    \hdashline
    \multirow{1}[2]{*}{Relevance} & L
    & .640 & .474
    & .510 & .450
    & .746  & .476
    & .569  & .497
    & .395  & .479
    & .328  & .480
    & .434  & .502
    & .369  & .508      \bigstrut[t]\\
          
    \multirow{1}[2]{*}{CAM}& P
    & .849 & .430
    & .765  & .389
    & .834  & .470
    & .690  & .478
    & .616  & .458
    & .570  & .440
    & .574  & .498 
    & .521  & .503 \bigstrut[b]\\
    \hdashline
    
    \multirow{2}[2]{*}{RSP}  & L
    & .849
    & .364
    & .712
    & .363
    
    & .859
    & .431
    & .749
    & .415
    
    & .540
    & .343
    & .479
    & .348
    
    & .558
    & .441
    & .504
    & .421
    \bigstrut[t]\\
    
      & P
    & .946
    & .348
    & .903
    & .324
    
    & .909
    & .437
    & .836
    & .410
    
    & .725
    & .337
    & .680
    & .331
    
    & .688
    & .438
    & .654
    & .431
    \bigstrut[t] \\
    \hdashline
    \multirow{2}[2]{*}{AGF}  & L
    & .824
    & .279
    & .672
    & .274
    
    & .645
    & .364
    & .507
    & .369
    
    & .492
    & .286
    & .430
    & .283
    
    & .395
    & .371
    & .353
    & .370
    \bigstrut[t]\\
    
      & P
    & .925
    & .268
    & .882
    & .255
    
    & .717
    & .368
    & .679
    & .364
    
    & .703
    & .281
    & .661
    & .269
    
    & .553
    & .367
    & .543
    & .362
    \bigstrut[t] \\
    \hdashline
    \rowcolor[rgb]{ .906,  .902,  .902}     \multirow{2}[2]{*}{AE}  & L
    & \textbf{.878}
    & \textbf{.125}
    & \textbf{.758}
    & \textbf{.118}
    
    & \textbf{.897}
    & \textbf{.191}
    & \textbf{.797}
    & \textbf{.178}
    
    & \textbf{.553}
    & \textbf{.124}
    & \textbf{.498}
    & \textbf{.122}
    
    & \textbf{.578}
    & \textbf{.187}
    & \textbf{.527}
    & \textbf{.183}
    \bigstrut[t]\\
    
    \rowcolor[rgb]{ .906,  .902,  .902} 
    \multirow{-2}{*}{AE} & P
    & \textbf{.958}
    & \textbf{.124}
    & \textbf{.923}
    & \textbf{.117}
    
    & \textbf{.961}
    & \textbf{.197}
    & \textbf{.937}
    & \textbf{.184}
    
    & \textbf{.747}
    & \textbf{.124}
    & \textbf{.707}
    & \textbf{.120}
    
    & \textbf{.708}
    & \textbf{.195}
    & \textbf{.674}
    & \textbf{.192}
     \bigstrut[t] \\
    \specialrule{2.0pt}{1pt}{1pt}
    \end{tabular}%
    }
    \caption{Performance of pointing game on Pascal VOC 2007 test set and COCO 2014 validation set. Each method reports different cases of each step for testing: $P$: Only predicted classes, $L$: All labels. ALL and DIF represent full data and subset of difficult images, respectively. PR represents positive ratio of attributions with respect to entire size of input image. -- denotes those attributions in pixels with all positive.} \label{table:tab1}
\end{table*}%
\subsection{Qualitative Assessment}
\subsubsection{Illustrative Example}
To qualitatively evaluate the attributions obtained from each approach, we examine the visual differences to compare how pixels with high relevance scores are clustered in the target object. The consistency of positive relevance among the approaches can be compared since all attribution methods aim to identify the most relevant components. In the qualitative evaluation, which relies heavily on human judgment, we focus on the following comparisons: i) class-specific attributions: whether the visualization accurately represents the desired class, ii) detailed descriptions of neuron activations that align with human perception and intuitive understanding, and iii) fine-grained attributions and objectness, which are central to the rationale for a decision, minimizing false positives in irrelevant areas.
\begin{figure}[t!]
  \centering
  \includegraphics[width=1\linewidth]{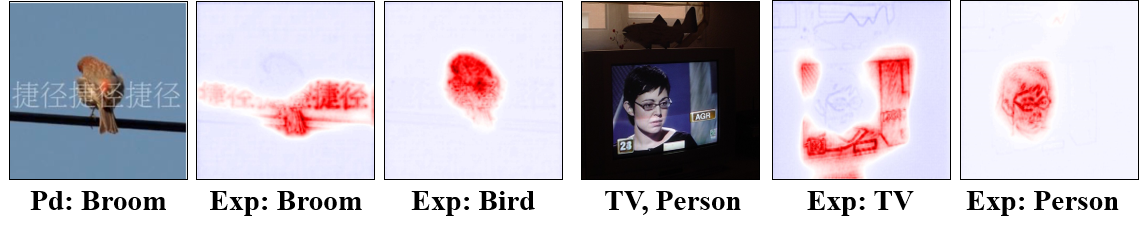}
    \caption{Left: Misclassification of the DNN due to Watermarks and corresponding explanations; Right: Discernment of edges within the image.}
  \label{fig:water}
\end{figure}
Fig.~\ref{fig:comp} shows comparison of Grad-CAM, Composite-LRP (LRP-COMP), contrastive excitation backprop (c*EB), contrastive LRP (c*LRP), attribution guided factorization (AGF), relevance CAM, and relative sectional propagation (RSP) results with the predictions of VGG-16 on the Pascal dataset. To clarify class-agnostic interpretability, we select the images containing two clear objects and propagate the attribution from each label. To verify the robustness and generalization, Fig.~\ref{fig:model} illustrates the heatmaps of our method for the output predictions with various models---AlexNet, InceptionNet, and DenseNet---on the ImageNet validation dataset. All attributions in the visualization are derived from the propagation rule in Eq.~\ref{eq:9} with $\gamma=1$. Compared with the attributions from other methods, those from our method are distributed in the key parts of the target objects (e.g., face and wheel), with clear separations between the target and other objects (including the background). These result in fine-grained visualization of evidence to decision, consistent with our assumptions.
\begin{figure}[t!]
  \centering
  \includegraphics[width=1\linewidth]{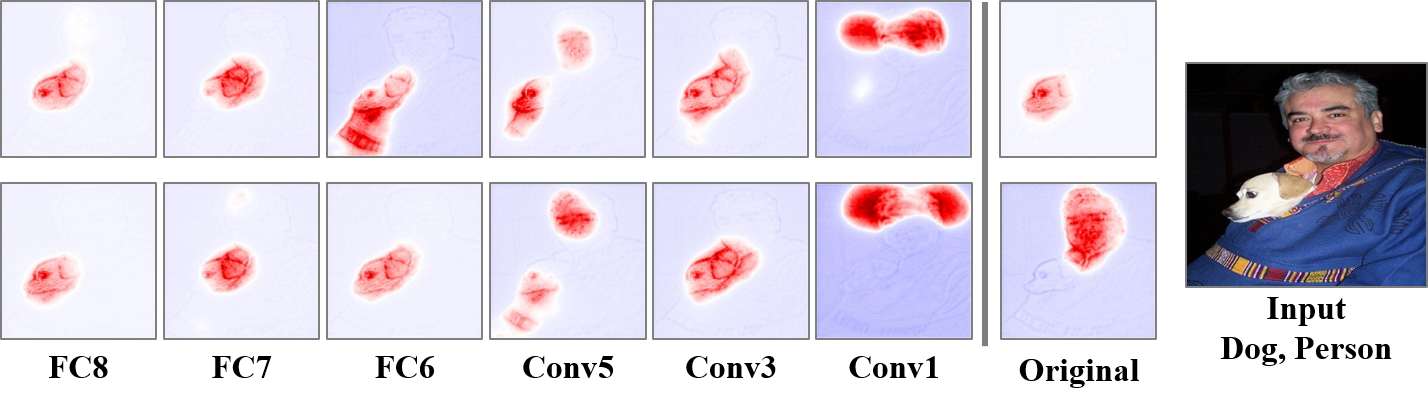}
    \caption{Sanity check: applying gradual initialization of VGG-16 model weights from end to beginning layer(First row: Dog, Second row: Person).}
  \label{fig:sal}
\end{figure}
Additionally, external factors, e.g. watermarks, can cause the DNN to make incorrect decisions. In Fig.~\ref{fig:water}, the image is misclassified by the DNN as a ``Broom" due to the watermark. In the next image, the rationale behind the misclassification is clearly visualized by our method. Also, when explaining the original bird class, we can confirm that the DNN still recognizes the bird information. The second image in Fig.~\ref{fig:water} demonstrates discernment in highlighting the main object while avoiding irrelevant edges. Some methods in Fig.~\ref{fig:comp} emphasize irrelevant edges or the borders of images indiscriminately. This might occur because activated neurons corresponding to the edges in the initial feature extraction stages have the potential to receive positive relevance during the backward propagation. Through Uniform shifting (Eqs.~\ref{eq:9}--~\ref{eq:10}), we ensure that edges contributing to contradictory evidence continue to receive negative relevance, thereby demonstrating discernment between irrelevant edges and the key object. 

\subsubsection{Sanity Check}
Sanity check\cite{adebayo2018sanity} deals with the insensitivity problem of some attribution methods when the parameters of the DNN are initialized in a cascading manner. It is critical to demonstrate that our explanation and the model decision are mutually reliant. Fig.~\ref{fig:sal} shows the variations in the attributions with the progressive randomization of certain layer weights. The attributions from each label are severely altered by the distortion of the parameters. 
\begin{figure}[t!]
  \centering
  \subfloat{\includegraphics[width=1\linewidth]{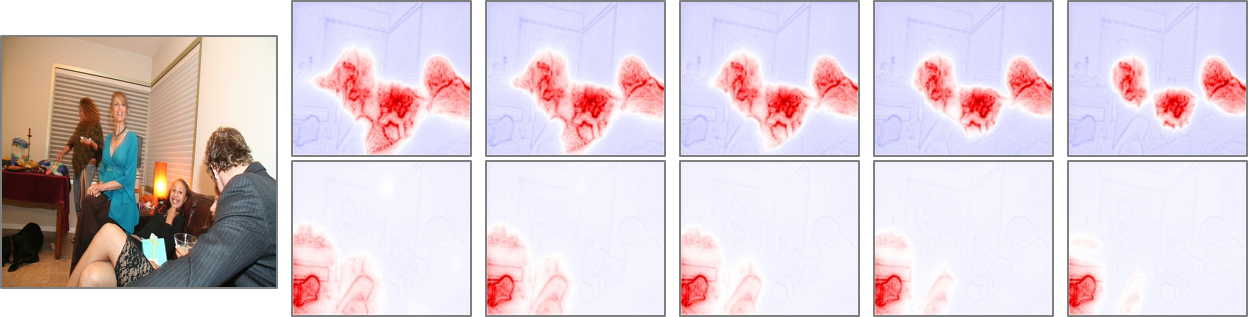}}\\
  \subfloat{%
  \small
    \begin{tabular}{cccccc}
    \specialrule{1.3pt}{1pt}{1pt}
    Model & $\gamma=2$ & $\gamma=1.75$ & $\gamma=1.5$ & $\gamma=1.25$ & $\gamma=1$ \\
    \midrule
    VGGNet & 87.2 & 87.4 & 87.6 & 87.6 & \textbf{87.8} \\
    ResNet & 88.5 & 88.6 & 88.8 & 89.1 & \textbf{89.7} \\
    \specialrule{1.3pt}{1pt}{1pt}
    \end{tabular}%
  }
  \caption{Variation in performance of pointing game with parameter $\gamma$. Evidence for decision can be increased or decreased by user preferences.}
  \label{fig:fig-tab}
\end{figure}
\begin{table}
\centering
\setlength{\tabcolsep}{10pt} 
\begin{tabular}{cccc|cc}
\specialrule{1.4pt}{1pt}{1pt}
$\mathbf{C}$ & $\mathbf{A}$ & $\mathbf{U}$ & $\Psi$ &  VGG & ResNet\\
\hline
\cmark & - & - & - & 83.9 / 69.6 & 85.6 / 72.6\\
\cmark & - & - & \cmark & 87.2 / 73.2 & 88.5 / 76.9 \\
- & \cmark & \cmark & - & 84.3 / 71.8 & 81.2 / 68.4 \\
\cmark & \cmark & \cmark & - & 83.3 / 71.1 & 86.4 / 74.1 \\
\hline
\cmark & \cmark & \cmark & \cmark & \textbf{87.2 / 74.8} & \textbf{88.5 / 77.1} \\
\specialrule{1.4pt}{1pt}{1pt}
\end{tabular}
\caption{Ablation study on variants of propagation rule in Eq.\ref{eq:9}. Each row presents performance of pointing game (ALL/DIF) based on different components. To prevent divergence due to Eq.\ref{eq:3}, $\gamma$ is set as 2.} \label{table:abl}
\end{table}
\subsection{Quality Evaluations of Attributions}
\subsubsection{Localization}
Pointing game \cite{zhang2018top} is a metric for evaluating attributions by determining the matching scores of the localization between the highest relevance point and the semantic annotations, e.g., the bounding box, of the object categories in an image. For each category, the localization accuracy is calculated as $Acc=\frac{\#Hit}{\#Hit+\#Miss}$, where $\#Hit$ scores if the pixel with the highest relevance score is within the object. However, because attributions are inevitably affected by the predictive performance of the model, decomposition on labels not identified by the model could lead to misinterpretation. Therefore, we assess both cases: (i) P: from predictions and (ii) L: all labels, and the results are listed in Tab.\ref{table:tab1}. In addition, we report the ratio of the pixels with positive relevance scores, notated as PR, to the total numbers of pixels in the image. Methods with the notation c* apply contrastive variation rules to set comparative classes for resolving the class-agnostic issue. As shown in the table, our method substantially outperforms other attribution methods in finding the target object in both cases, despite occupying extremely few positive spaces.

Fig.~\ref{fig:fig-tab} shows the variations in the pointing game performance with parameter $\gamma$. Recall, this parameter denotes the user preference that defines the allocation amount of positive evidence. As the parameter decreases, the allocation of attributions is expected to be concentrated in a tighter area, enhancing localization performance by removing contradictory evidence.
As an ablation study, we examine different combinations of the propagation rule in Eq.~\ref{eq:9}, with results summarized in Tab.~\ref{table:abl}. The group of $\mathbf{C}$ and $\Psi$ represents our base concept, maintaining the evidence for a decision while converting elements corresponding to the opposite position, such as irrelevant objects and the background, to negative values. $\mathbf{A}$ and $\mathbf{U}$ enhance propagation robustness by reducing sensitivity to locally activated features in the shallow layers. 

\begin{figure}[t!]
  \centering
  \subfloat{\includegraphics[width=1\linewidth]{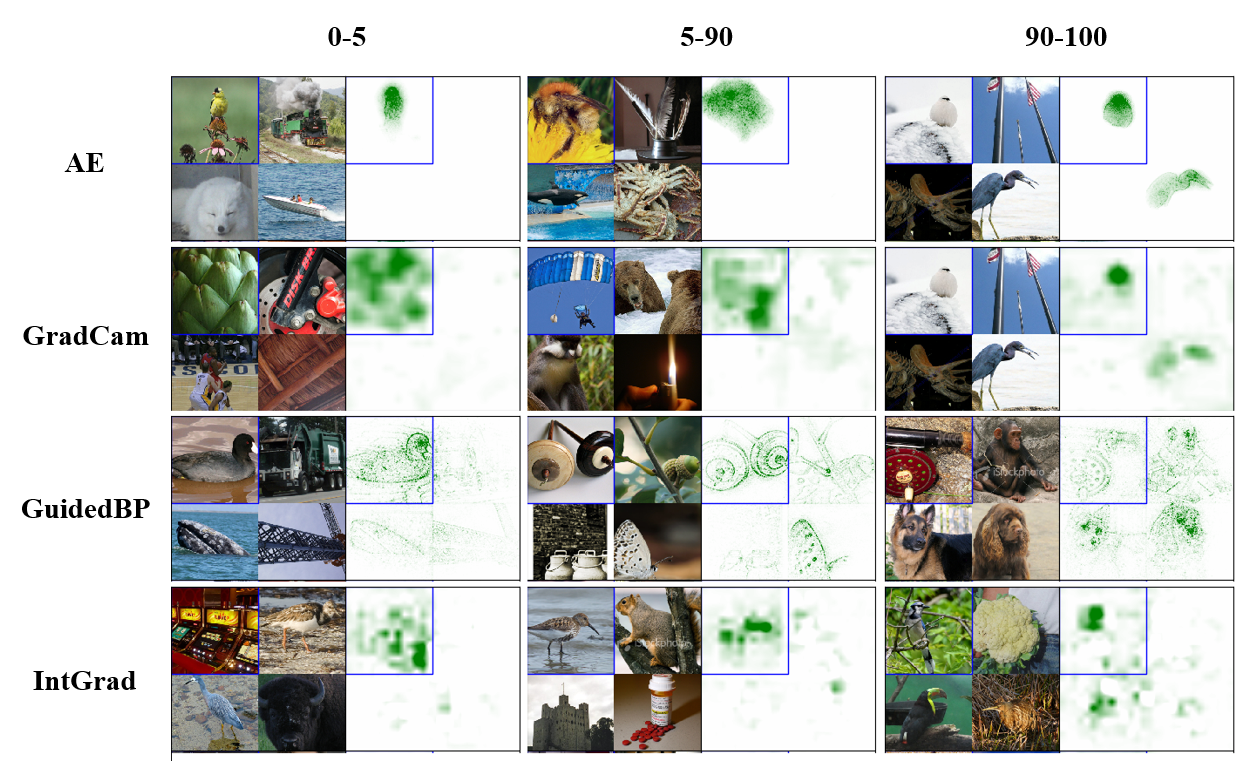}}\\
  \subfloat{%
  \small
    \begin{tabular}{ccccc}
    \specialrule{1.3pt}{1pt}{1pt}
     Model& G-CAM & Guid-BP & IntGrad & \textbf{Ours}\\
    \midrule
    VGGNet & 72.5 (87.5) & 38.9 (45.3) & 72.3 (76.7) & 93.8\\
    ResNet & 81.6 (97.7) & 39.3 (47.5) & 81.7 (97.7) & 99.4\\
    \specialrule{1.3pt}{1pt}{1pt}
    \end{tabular}%
  }
  \caption{Visual results (VGGNet) and performance Comparison Using GridPG. Images and attributions are sorted in descending order of localization. The values in parentheses represent the thresholds applied based on the ratio of positive attributions yielded by our method.}
  \label{fig:fig-grid}
\end{figure}

Furthermore, we employ the GridPG\cite{Rao2022CVPR} as an additional performance evaluation to assess the localization capability of attribution techniques, extending the Pointing Game. We adhere to the experimental settings used in the original paper to maintain generality in our evaluation and report both qualitative and quantitative evaluations in Fig \ref{fig:fig-grid}. We benchmark our method against prominent attribution approaches: Grad-CAM, Guided Backprop, and Integrated Gradients. For each methodology, we selected layers [input, middle, final] that demonstrated optimal performance as reported in the paper\cite{Rao2022CVPR}. Our method selects the input layer for visualization, fully decomposing from the output to the input layer. Additionally, for a fair evaluation, we apply top-k thresholding to the results of other methods, represented in parentheses. The thresholding value is derived from the ratio of positive attributions in our results relative to the total image size. This process assesses how effectively each method localizes the essential input features by extracting attributions in descending order of their values. Our method demonstrates significantly superior localization performance compared to other approaches. Each bin at the top represents the descending order of localization performance, and each row displays representative images within that performance range. As seen in the figure, our method accurately identifies the main regions of the images. Even in the lower localization ranges, the observed effects can be attributed to class overlaps within the images.

\begin{table}[t]
\RawFloats
\centering
\resizebox{1\columnwidth}{!}{
\begin{tabular}{ccccccccc}
\specialrule{2.0pt}{1pt}{1pt}
    Model      &     Metric    & c*LRP &  c*EB   &RSP  & RCAM &AGF & \textbf{Ours}   \\
\midrule
\multirow{2}{*}{VGGNet}    & mAP    
& 0.356  & 0.353 & \textbf{0.603} & 0.393 & 0.598 & 0.549\\
          & Out/In 
& 0.979 & 0.983 & 0.670  & 0.965 &  0.678 & \textbf{0.456} \\
\midrule
\multirow{2}{*}{ResNet} & mAP    
& 0.516  & 0.428 &  0.608 & 0.505 & \textbf{0.622} & 0.559\\
          & Out/In 
& 0.926 & 0.883 & 0.785 & 0.682  & 0.767  & \textbf{0.490}\\
\specialrule{2.0pt}{1pt}{1pt}
\end{tabular}
}
\caption{Performance of mAP and outside--inside ratio over ImageNet segmentation dataset. Attribution methods that have class-discriminativeness are compared. Low Out/In ratio implies attributions are intensively distributed in semantic mask.} \label{table:tab2}
\end{table}
\begin{figure*}[t!]
  \centering
  \includegraphics[width=1\linewidth]{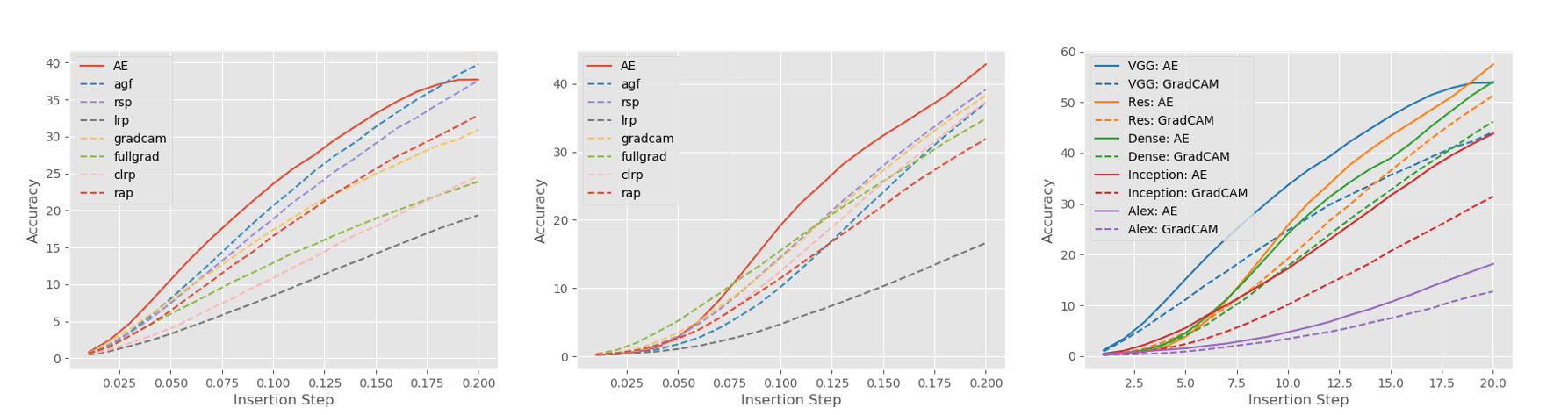}
    \caption{Variation in model accuracy with incremental MoRF insertion: from 1\% to 20\% total pixels in increments of 1\%. First and second graphs compare attribution methods on VGG-16 and ResNet-50, respectively. Third graph compares our method and baseline (Grad-cam) on various models.}
  \label{fig:graph}
\end{figure*}

\subsubsection{Objectness of Attributions}
In the field of weakly supervised segmentation (image-label level supervision), many studies \cite{ahn2019weakly, melas2022deep, huang2018weakly, chen2022class} have obtained the initial seeds by interpreting predictions that include broad localization information. Although assigning attributions and segmentation can be closely related in seeking pixels corresponding to a predicted object, it is unlikely that all areas of the object have equally significant effects on the decision. Evaluating the distribution of attributions in salient parts of an object is challenging because there is no ground truth for the importance of annotations. However, some inferences can be made from the density of the relevance scores.

In this regard, \cite{lapuschkin2016analyzing} introduced a metric called the outside--inside ratio, to evaluate the concentration of the relevance on the target object by comparing the relevance scores in/outside of the annotations, e.g., the bounding box and the segmentation mask. We use this method to determine the degree to which attributions are centrally distributed within the mask of an object. We mainly compare the conventional methods that deal with the class-agnostic issue and provide detail descriptions of the input features.
\begin{equation}
    \mu = \frac{\frac{1}{\left\vert P_{out} \right\vert}\sum_{q\in P_{out}}N(R_q^{(0)})}
    {\frac{1}{\left\vert P_{in} \right\vert}\sum_{p\in P_{in}}N(R_p^{(0)})}.
\end{equation}
Here, \(\left\vert \cdot \right\vert\) denotes the cardinality operator, which measures the size of each set, inside ($P_{in}$) and outside ($P_{out}$) of the mask. $N$ represents normalization in the scope of $[0,1]$. Because the compared methods have both positive and negative relevance scores and there can be cases in which there is no positive value inside the mask, all values can show divergence. To prevent this, we add normalization to the original metric, changing all relevance scores to positive while maintaining the degree of contribution.

We utilize the ImageNet segmentation dataset ~\cite{Guillaumin2014ImageNetAW}, which consists of 4,276 images with segmentation masks. Tab.\ref{table:tab2} compares different attribution methods---c*LRP, c*EB, RSP, RCAM, and AGF---based on the metric mean average precision (mAP) and the outside--inside ratio. Because positive relevance is distributed out of the segmentation mask, the value of \(\mu\) is increased. By contrast, when the high-priority attributions are accumulated in in the mask, the ratio is decreased. 
Based on Tab.~\ref{table:tab2}, RSP and AGF show impressive segmentation performance without any additional supervision. In comparison, our method performs slightly lower based on the mAP, whereas it shows a superior performance than the other methods in terms of the outside--inside ratio metric, indicating that high-priority attributions are intensively distributed inside the segmentation mask, which are mainly significant input features related to the decision.

\subsection{Analysis of Perturbation Metric}
\begin{figure}[t!]
  \centering
  \includegraphics[width=1\linewidth]{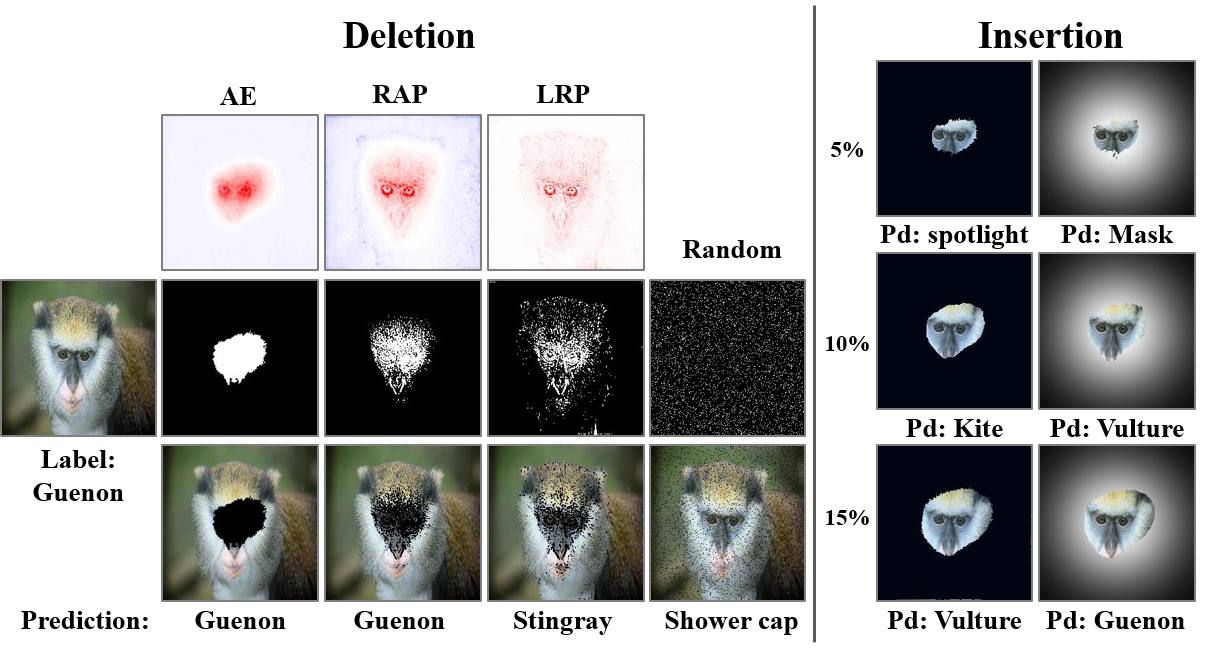}
    \caption{Examples for addressing issues of region perturbation metrics. Left side represents the case of vulnerability to perturbation. Alternatively, we evaluate the increase in accuracy with the rise of high-relevance pixels against a background of a Gaussian filter. Detailed explanation is described in Section 5.4.}
  \label{fig:perturb}
\end{figure}

Intuitively, when evaluating assigned attributions, it is believed that removing/inserting high-relevance pixels should significantly reduce/increase the prediction accuracy. \cite{samek2017evaluating} introduced a method for quantitatively assessing explanations methods using the region perturbation process, starting with the most relevant first (MoRF) and the least relevant first (LeRF). They formalized this method as an area over the perturbation curve. Similarly, \cite{petsiuk2018rise} proposed Insertion/Deletion score metric that measures an increase or decrease in prediction probability when important pixels are inserted or removed. However, in empirical experiments, we found that the prediction of a DNN is more distorted by scattered noise than by fine perturbation in detailed region. This issue is related to the field of adversarial attacks \cite{goodfellow2014explaining}, which addresses the vulnerability of a DNN to an adversarial perturbation. Fig.~\ref{fig:perturb} shows samples illustrating perturbation issues. An image containing ``guenon'' is correctly classified with the VGG-16 model. In the figure, the first row shows the attributions from each method, and the second row represents the mask corresponding to the top 4,000 pixels in order of relevance value from each method. The third row shows the perturbed images in which the area corresponding to the mask is erased. RAP and AE produce attribution maps clustered in the center of the face, and the predicted classes show no changes when the mask is removed. Conversely, LRP and random noise, particularly the latter which is not interpretable from a human-view, change the prediction results to ``Stingray'' and ``Shower cap'', respectively. 
On the other hand, we empirically found that the insertion-based evaluation metric reflects the prediction score more stably and aligns with the intended experimental design. However, inserting areas into blank spaces can lead to another misconception due to the abrupt change in pixel values at the boundaries of the inserted images. To enhance stability, we rebuilt the Imagenet validation dataset by extracting the pixels corresponding to the MoRF based on a predetermined percentage and inserting them into a blank image with a Gaussian filter applied. The right image of Fig.~\ref{fig:perturb} presents an example of the performance evaluation method through insertion. Adding a simple Gaussian filter helps mitigate the gap at the boundaries, thereby increasing the stability of the metric.

The first and second graphs in Fig.~\ref{fig:graph} show the results of applying the MoRF insertion with a Gaussian filter on VGG-16 and ResNet-50, respectively. The third graph presents the results of various models: VGG-16, ResNet-50, DenseNet-121, Inception-V3, and AlexNet, with the baseline result of Grad-CAM. For a reasonable comparison in the third graph, the accuracy of each model is divided by the original model performance. For each step, we perturb 1\% of pixels corresponding to the MoRF, totaling 20\% pixel distortion per image in increments of 1\%. Based on the results, our method shows a rapid increase in accuracy compared to existing methods. Thus, our method is the best at finding the most essential parts of the input as a condensed region with a limited positive relevance ratio.
\section{Discussions}
To our best knowledge, there is still no perfect elucidation of the complex inner mechanisms of DNNs, excluding their structural and conceptual design. Therefore, it is crucial to provide the best explanations of a decision due to the possibility of unexpected phenomena such as ``clever Hans\cite{lapuschkin-ncomm19}''. Explanation methods that show a high degree of approximation in terms of objectness for a network offer many potential advantages in increasing the model confidence. However, providing ``visually pleasing'' explanations that are overly appealing to users carries its own set of risks. For example, in the sanity check illustrated in Fig.~\ref{fig:sal}, distortions in the model weights lead to incorrect explanations, yet the end result still appears visually plausible. The algorithm is designed to emphasize the highly contributed areas, but this can also produce plausible explanations for imperfect models. Therefore, it is advisable not to rely solely on one method of explanation but to use a combination of approaches, and it is essential to accompany these models and their explanations with thorough performance evaluations to ensure their validity. Additionally, the evaluation of explainability still relies on human subjectivity, and as demonstrated by the unexpected issues in Section 5.4, there remain many areas for improvement.
\section{Conclusion}
In this paper, we introduced a novel method for interpreting network predictions called Attribution Equilibrium. Our method decomposes output predictions into essential attributions, balancing positive and negative relevance for clearer visualization of the evidence behind network decisions. We have shown that this approach maintains the overall importance of evidence, avoiding conflicts between positive and negative attributions, and providing a more sophisticated and fine-grained visualization of relevant input features. We evaluated our proposed method both quantitatively and qualitatively to confirm the quality of the attributions. The results demonstrate that the attributions from our method provide clearer visualizations of key input features, as well as class-specific and detailed descriptions of neuron activation. This approach enhances the interpretability of neural network decisions, making it a valuable tool for understanding the underlying mechanisms of deep learning models.

\ifCLASSOPTIONcompsoc
  \section*{Acknowledgments}
\else
  \section*{Acknowledgment}
\fi

This work was supported by the National Research Foundation of Korea(NRF) grant funded by the Korea government(MSIT) (No. RS-2024-00449891) and was partly supported by BK21 FOUR project (AI-driven Convergence Software Education Research Program) funded by the Ministry of Education, School of Computer Science and Engineering, Kyungpook National University, Korea (41202420214871) and Institute of Information \& Communications Technology Planning \& Evaluation(IITP) grant, funded by the Korea government(MSIT) (No. RS-2019-II190079, Artificial Intelligence Graduate School Program (Korea University) and No. RS-2022-II220984, Development of Artificial Intelligence Technology for Personalized Plug-and-Play Explanation and Verification of Explanation).

\ifCLASSOPTIONcaptionsoff
  \newpage
\fi



%
\bibliographystyle{IEEEtran}
\bibliography{egbib}
%



%

\begin{IEEEbiography}[{\includegraphics[width=1in,height=1.25in,clip,keepaspectratio]{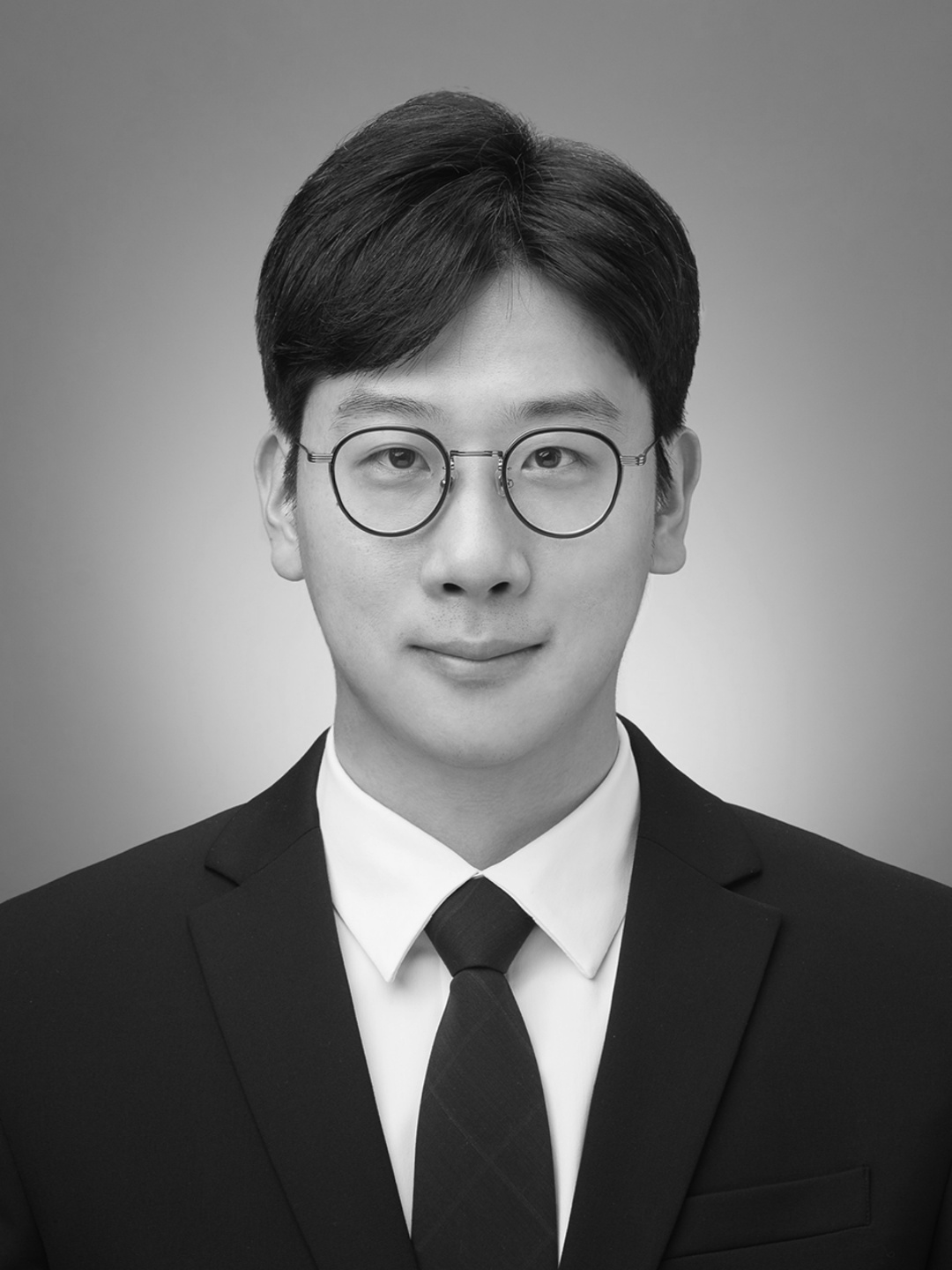}}]{Woo-Jeoung Nam}
    received the B.S. degree in computer science and engineering from Hanyang University, Seoul, South Korea, in 2016, and an integrated master's and Ph.D. degree from the Department of Computer and Radio Communications Engineering, Korea University, Seoul in 2022. He is currently an Assistant Professor with the School of Computer Science and Engineering, Kyungpook National University (KNU). His current research interests include machine learning, computer vision and explainable artificial intelligence.
\end{IEEEbiography}


\begin{IEEEbiography}[{\includegraphics[width=1in,height=1.25in,clip,keepaspectratio]{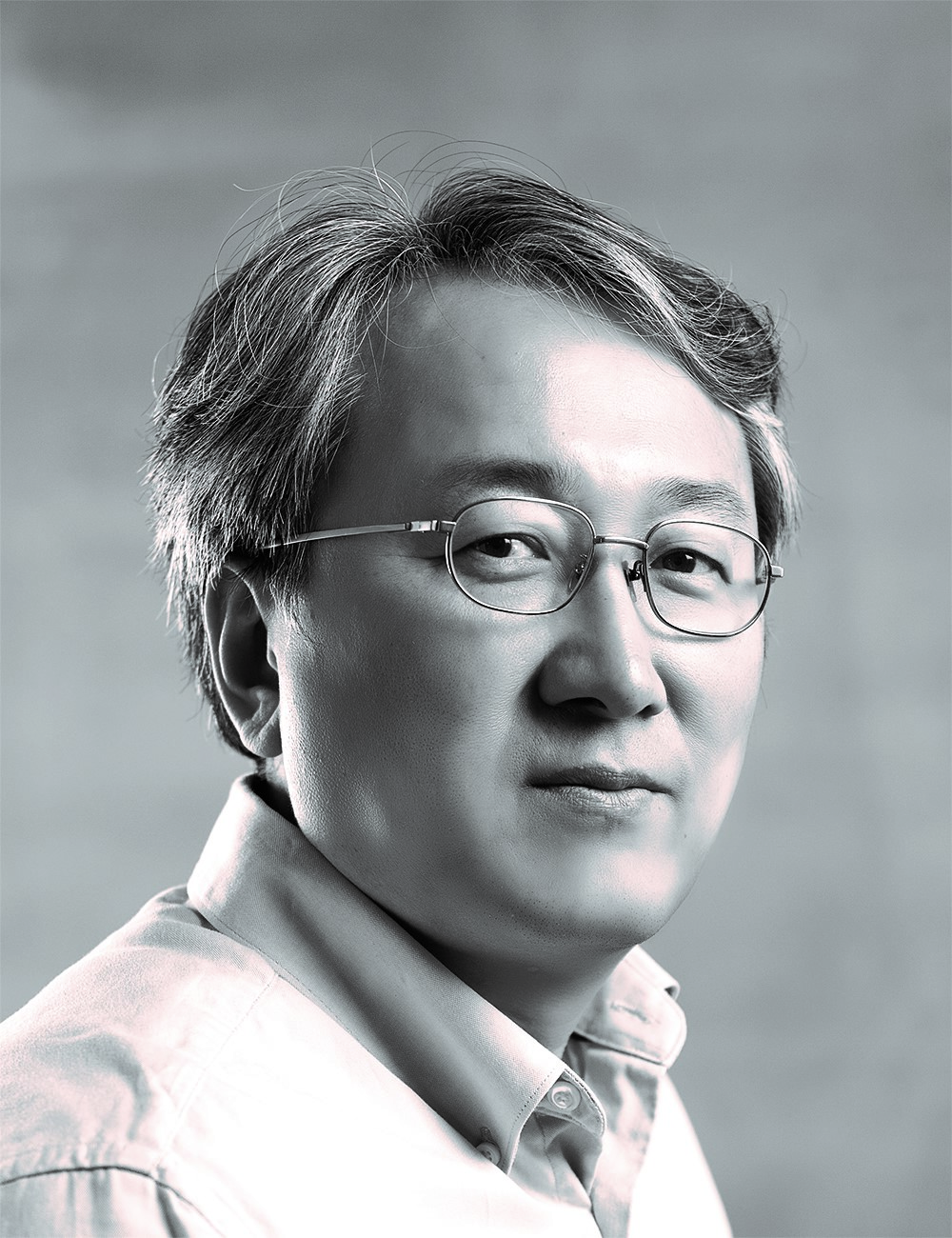}}]{Seong-Whan Lee}
    received the B.S. degree in computer science and statistics from Seoul National University, Seoul, South Korea, in 1984, and the M.S. and Ph.D. degrees in computer science from the Korea Advanced Institute of Science and Technology, Daejeon, South Korea, in 1986 and 1989, respectively. He is currently the Head of the Department of Artificial Intelligence, Korea University, Seoul, South Korea. He is also a Fellow of the International Association of Pattern Recognition (IAPR) and the Korea Academy of Science and Technology. His research interests include artificial intelligence, pattern recognition, and brain engineering.
\end{IEEEbiography}







\end{document}